%% file: main.tex
\documentclass{article}

\usepackage[table]{xcolor}
\usepackage[preprint]{corl_2026}
\usepackage{booktabs}
\usepackage{graphicx}
\usepackage{makecell}
\usepackage{multirow}
\usepackage{capt-of}
\usepackage{listings}
\usepackage{float}
\usepackage{tikz}
\usetikzlibrary{arrows.meta,positioning}
\usepackage[T1]{fontenc}

\newcommand{\best}[1]{\textbf{#1}}
\definecolor{corrblue}{HTML}{2B6CB0}
\definecolor{corrorange}{HTML}{D9480F}
\newcommand{\corrneg}[2]{\cellcolor{corrblue!#1}#2}
\newcommand{\corrpos}[2]{\cellcolor{corrorange!#1}#2}

\definecolor{promptvar}{HTML}{D9480F}
\definecolor{promptschema}{HTML}{2B6CB0}

\lstdefinestyle{prompttemplate}{
  basicstyle=\scriptsize\ttfamily,
  frame=single,
  breaklines=true,
  columns=fullflexible,
  keepspaces=true,
  moredelim=**[is][\color{promptvar}]{<<VAR>>}{<</VAR>>}
}

\title{Critical Interval MSE: Toward Reliable Offline Validation for Robot Manipulation Policies}

\author{
  \textbf{Haoxu Huang$^{1,2}$} \quad 
  \textbf{Tongsam Zheng$^{1}$} \quad 
  \textbf{Yifan Chen$^{1}$} \quad 
  \textbf{Jiacheng You$^{1,2}$} \quad 
  \textbf{Yang Gao$^{1,2}${\footnotemark[1]\thanks{Corresponding author}}} \\
  \\
  \textsuperscript{1 }Tsinghua University \;\;
  \textsuperscript{2 }Shanghai Qi Zhi Institute\;\;\\[0.3cm]
}

\lstdefinestyle{skilldoc}{
  basicstyle=\scriptsize\ttfamily,
  frame=single,
  aboveskip=\baselineskip,
  belowskip=0.8\baselineskip,
  breaklines=true,
  columns=fullflexible,
  keepspaces=true,
  showstringspaces=false,
  upquote=true
}
\begin{document}
\hypersetup{pdfauthor={Haoxu Huang, Tongsam Zheng, Yifan Chen, Jiacheng You, Yang Gao}, pdfsubject={}}
\maketitle

\vspace{-7mm}
\input{sections/abstract}


\input{sections/introduction}
\input{sections/related_work}

\input{sections/method}

\input{sections/experiments}

\input{sections/conclusion}

\input{sections/acknowledgments}

\bibliography{references}

\clearpage
\appendix
\input{sections/appendix}

\end{document}

%% file: sections/abstract.tex
\begin{abstract}
Real-world evaluation is the gold standard for robot manipulation policies because it tests them against the physical conditions and deployment challenges they are ultimately designed to handle. However, real-world evaluation is also the bottleneck for iterating on robot imitation learning policies: it is costly, difficult to reproduce, and often too sparse to reliably compare nearby model variants. A straightforward proxy for performance is validation loss on expert demonstrations, but this proxy is often poorly correlated with real-world performance. In this paper, we introduce Critical Interval MSE (CI-MSE), an intuitively simple yet effective offline validation metric. CI-MSE uses a zero-shot agent to localize task-critical segments, restricts error computation to those segments, and applies simple action-alignment procedures that better match rollout-time behavior. Across simulation and real-world experiments, CI-MSE generally yields a stronger correlation between validation error and rollout performance than raw MSE. Across 27 distinct policy checkpoints, CI-MSE achieves a Spearman rank correlation of $-0.87$, closer to the ideal value of $-1$ than raw MSE's $-0.61$, demonstrating a significant improvement. Sensitivity analysis shows that our metric is robust across a wide range of hyperparameter settings. We further study the effectiveness of CI-MSE under evaluation distribution shifts and identify practical boundaries for its use. In summary, this paper provides a simple and reliable offline validation tool for accelerating policy iteration. Project webpage: \url{https://ci-mse.github.io/}
\end{abstract}

%% file: sections/introduction.tex
\section{Introduction}
\label{sec:introduction}

Rapid model iteration in robot imitation learning depends on having a validation signal that is inexpensive, reproducible, and predictive of real-world behavior. In practice, the gold standard remains policy rollouts on physical systems. However, such evaluations are expensive to run, difficult to standardize across laboratories, and often limited to a small number of trials~\cite{zhou2025autoeval,li2025simpler,atreya2025roboarena,li2025worldeval}. These constraints make it difficult to compare nearby policy variants, especially when researchers study minor changes in architecture, dataset size, or training recipes~\cite{khazatsky2024droid,openx2023rtx}.

\begin{figure}[t]
\centering
\includegraphics[width=\linewidth]{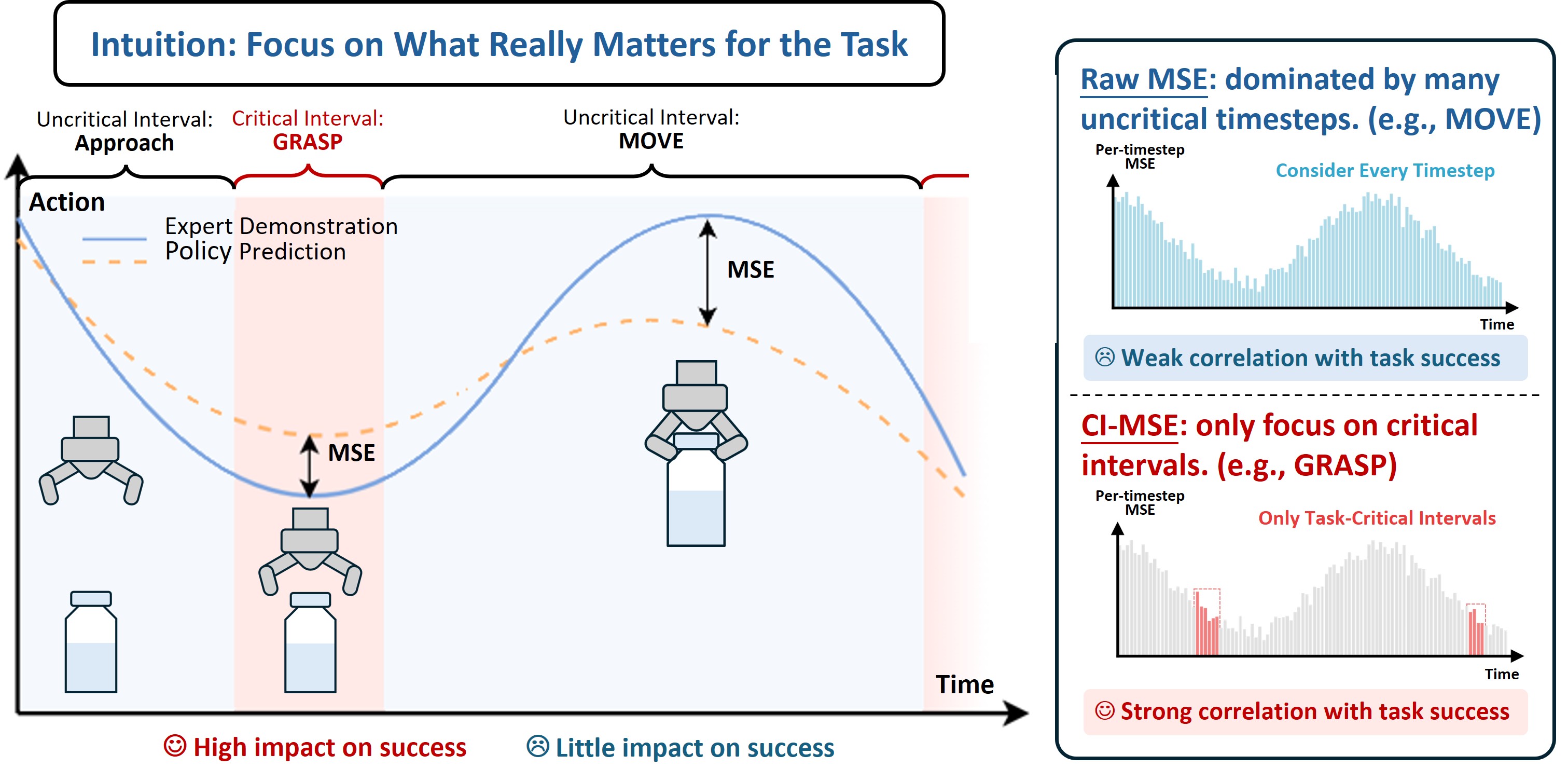}
\caption{Illustration of critical intervals in a robot manipulation trajectory. The example shows a robot gripper transferring a bottle. Aggregated MSE is dominated by noncritical motion, whereas critical interval MSE focuses on the critical grasp stage.}
\label{fig:intuition}
\end{figure}

Offline validation is attractive because it can be computed on held-out demonstrations without additional rollout time on robots~\cite{hussenot2021hyperparameter,mandlekar2022what}. Yet most practitioners agree that the common choice of action-space MSE often correlates weakly with actual task success~\cite{lin2025data,mandlekar2022what,florence2022implicit,pari2022surprising,bronars2026tune,tiezzi2025learning}. The mismatch arises partly because many timesteps in a trajectory are irrelevant to task completion but produce large action errors, whereas the truly consequential actions are brief, contact-rich, and sensitive to small errors~\cite{wen2021keyframe,johns2021coarse,tsuji2026survey}. Fig.~\ref{fig:intuition} illustrates this mismatch. Consider a robot gripper transferring a bottle. The grasp stage is closely related to task success, and the gripper must carefully align itself with the mouth of the bottle. In contrast, the transition stage is less relevant to task success: the gripper can move horizontally or lift the bottle higher before putting it down. However, relative to the expert trajectory, the validation error is disproportionately high during the transition stage because transition actions are diverse and occupy a large portion of the trajectory. Therefore, the average error is dominated by noncritical intervals, largely obscuring the performance signal. Moreover, rollout-time procedures such as temporal ensembling or real-time action chunking can materially alter policy behavior, even though they are usually ignored by offline metrics~\cite{zhao2023learning,black2026real}.

To address this problem, this paper makes three contributions. First, we introduce \emph{critical interval MSE} (CI-MSE), an offline validation metric that focuses on short task-critical intervals. Because noncritical actions dominate the validation error, an intuitively simple solution is to restrict error computation to task-critical intervals instead of averaging over entire demonstrations. We localize these intervals in a zero-shot manner using an agent skill that applies a fixed, task-independent episode protocol through VLM agents. In addition, we apply action-alignment methods, including temporal ensembling~\cite{zhao2023learning}, RTC~\cite{black2026real}, and dynamic time warping, to better match rollout-time behavior. Second, we assess the effectiveness of CI-MSE through both simulation and real-world experiments. We find that CI-MSE achieves a stronger validation--evaluation correlation than raw MSE. Across 27 model checkpoints that differ in architecture, dataset size, training steps, and VLM backbone, CI-MSE achieves a Spearman rank correlation of $-0.87$, whereas raw MSE achieves only $-0.61$. Real-world experiments on four tasks yield results consistent with those of the simulation experiments. Third, we evaluate the robustness of CI-MSE through systematic experiments. We test our metric under different training--evaluation distribution shifts, a common setting in evaluations of policy generalization~\cite{pumacay2024colosseum}. We find that although CI-MSE is affected by distribution shift, it still provides a more reliable ranking than raw MSE. Sensitivity analysis also shows that our metric is robust across a wide range of hyperparameter settings.

Our goal is not to claim that offline validation can replace gold-standard robot rollouts. Rather, we aim to make offline validation more useful for rapid model iteration while clarifying the design principles for building more reliable offline benchmarks using our proposed metric.

%% file: sections/related_work.tex
\section{Related Work}
\label{sec:related_work}

\textbf{Real-world robot policy evaluation.}
Real-world benchmarks provide the most faithful signal for policy quality, and recent work improves reproducibility through standardized robot setups, autonomous evaluation, physical skill suites, and distributed pairwise comparisons~\cite{zhou2023toto,zhou2025autoeval,chen2026manipulationnet,atreya2025roboarena}. However, physical rollouts remain expensive, hard to replicate across labs, and difficult to distinguish many nearby model variants, especially under limited trial budgets. 

\textbf{Simulation-based evaluation.}
Simulation benchmarks offer reproducibility, parallelism, and controlled distribution shifts, enabling large-scale evaluation of manipulation policies~\cite{james2020rlbench,gu2023maniskill2,liu2023libero,pumacay2024colosseum}. Notably, SIMPLER bridges the sim-to-real gap with careful parameter tuning and visual processing, and shows strong correlation with real-world performance across various policies~\cite{li2025simpler}. However, extending simulation to new tasks and scenarios is time-consuming and requires domain expertise.

\textbf{Offline evaluation and policy ranking.}
Compared with robot benchmark design, offline validation for robot policies is less established; related questions, however, have been studied extensively in RL and LLM evaluation.
Offline policy evaluation in RL estimates policy value from fixed logged data, using estimators or benchmarks designed for settings where new interaction is costly~\cite{jiang2016doubly,fu2021dope}. Since exact value prediction is often unnecessary for model selection, recent work also studies policy ranking directly~\cite{da2024popr,gu2024porank}. LLM evaluation has reached a related conclusion: scalable offline judges and pairwise comparisons are useful for ranking models, but must be checked against human-preference or task-level targets~\cite{zheng2023mtbench,tan2025judgebench}.

%% file: sections/method.tex
\section{Critical Interval MSE for Offline Validation}
\label{sec:method}

\begin{figure}[t]
\centering
\includegraphics[width=\linewidth]{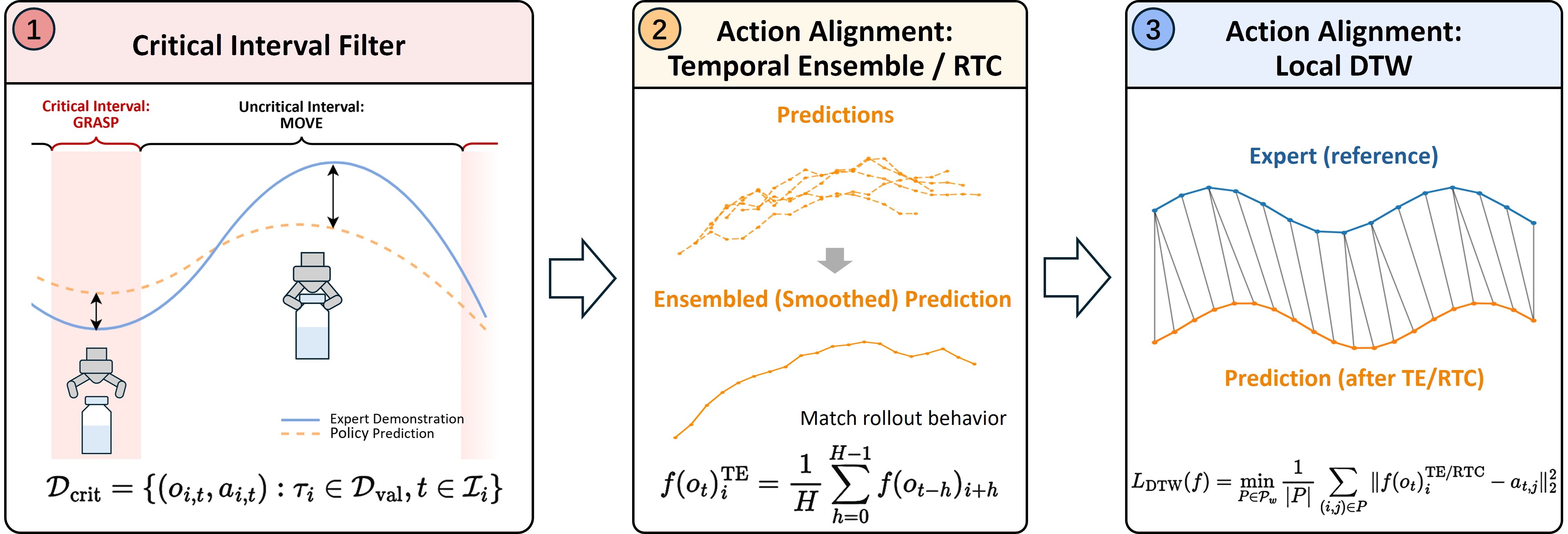}
\caption{Pipeline for critical interval MSE computation. A zero-shot agent skill localizes critical intervals using a fixed episode protocol. TE/RTC then matches rollout-time behavior, and DTW computes distance to expert actions on a locally minimizing warping path.}
\label{fig:pipeline}
\end{figure}

This section develops CI-MSE as an offline validation metric for ranking policies. We first formulate the desired agreement between validation error and rollout performance, then introduce two complementary components: a critical-interval filter that focuses on task-decisive timesteps and action alignment that matches rollout-time execution. Finally, we describe the Elo-based ordinal score used for real-world rollout evaluation. Fig.~\ref{fig:pipeline} shows the pipeline of CI-MSE computation.

\subsection{Problem Formulation}
\label{subsec:problem}

Let $f$ denote a robot manipulation policy, and let each validation trajectory be denoted by
\(\tau_i = \{(o_{i,t}, a_{i,t})\}_{t=1}^{T_i}\), where
\(o_{i,t}\) is the observation and \(a_{i,t}\) is the expert action.
The offline validation set is
\(\mathcal{D}_{\mathrm{val}} = \{\tau_i\}_{i=1}^{N}\). Let $r(f)$ denote a rollout-based evaluation score for policy $f$, such as task success rate or a partial progress score. We seek an offline metric $L(f;\mathcal{D}_{\mathrm{val}})$ whose induced ordering of policies agrees with the ordering under $r(f)$ as closely as possible. The metric $L$ aggregates timestep-level errors $\{\ell(f,o_{i,t}, a_{i,t}) : 1 \le i \le N, 1 \le t \le T_i\}$ over the validation set.

\begin{equation}
    \max_{L} corr\left(-L(f;\mathcal{D}_{\mathrm{val}}), r(f)\right)
\end{equation}

\subsection{Critical Interval Filter}
\label{subsec:cimse}

We define a \emph{critical interval} as a contiguous segment of a demonstration in which action accuracy has a significant impact on the task outcome. Typical examples include making object contact, grasping, inserting an object, and performing fine alignment near a task target. Actions in these intervals are subject to strict physical constraints and are therefore both sensitive to error and closely associated with success. By contrast, long transit motions or idle stabilization can contribute heavily to raw MSE without determining whether the task succeeds. Consequently, the aggregate error over the whole episode can be dominated by noncritical intervals and obscure the performance signal.

Given a set of critical intervals $\{\mathcal{I}_i\}_{i=1}^{N}$ on validation trajectories $\mathcal{D}_{\mathrm{val}}=\{\tau_i\}_{i=1}^{N}$, CI-MSE computes action error only over timesteps inside those intervals:
\begin{equation}
\mathcal{D}_{\mathrm{crit}} = \{(o_{i,t}, a_{i,t}) : \tau_i\in \mathcal{D}_{\mathrm{val}}, t \in \mathcal{I}_i\}
\end{equation}
In principle, we could give less weight to noncritical intervals and compute a weighted average error over the whole trajectory. However, tuning this weight complicates the metric and makes it much more fragile. Therefore, in this paper, we simply filter out noncritical timesteps. This straightforward change better aligns the metric with task structure by eliminating error contributions from uninformative timesteps.

\begin{figure}[t]
\centering
\includegraphics[height=0.32\textwidth]{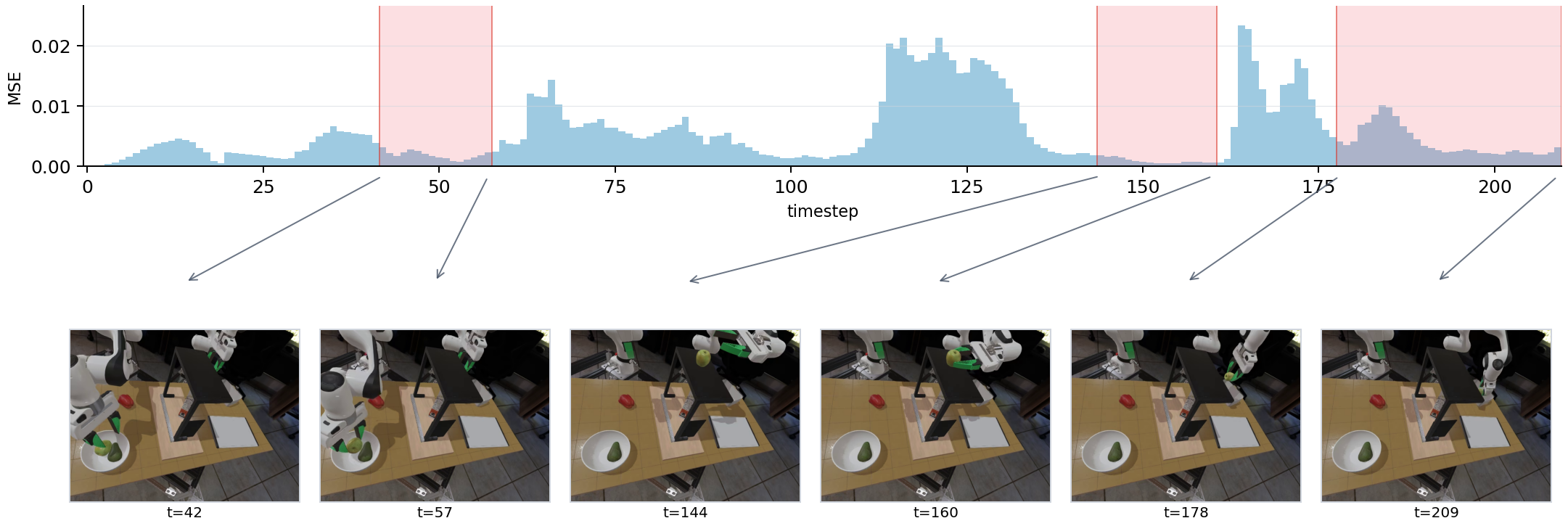}\hfill
\includegraphics[height=0.32\textwidth]{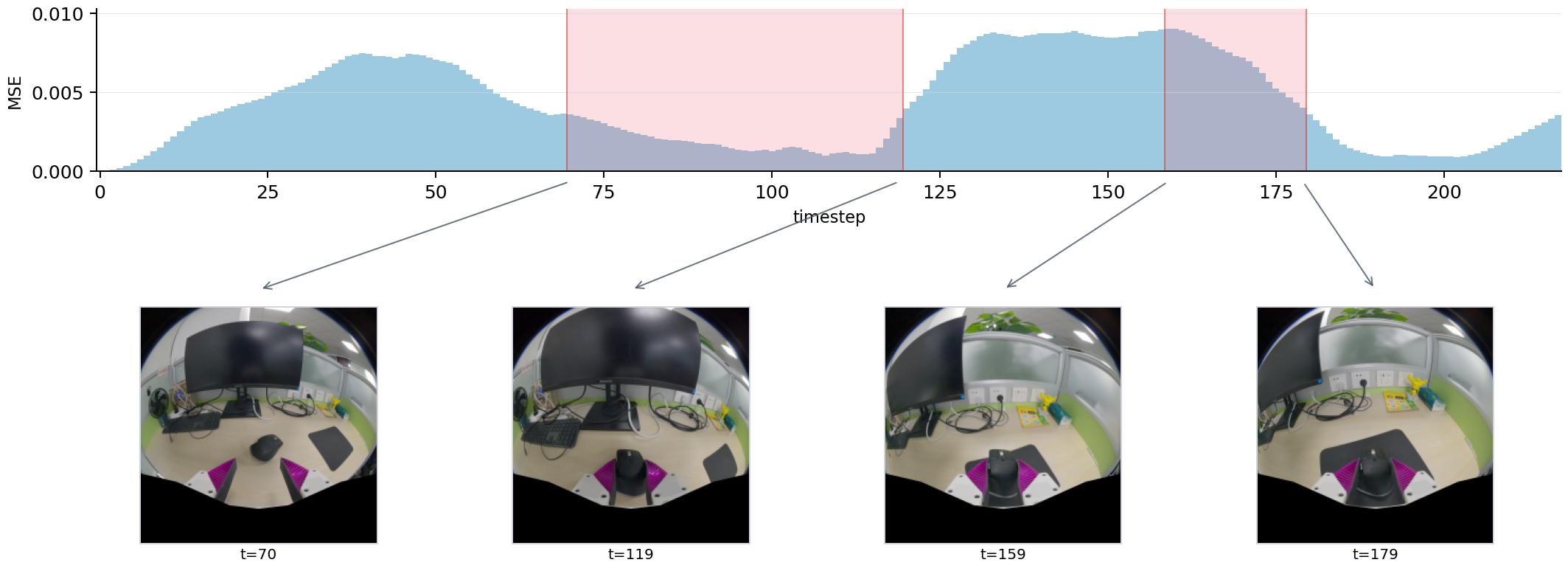}
\caption{Examples of zero-shot critical interval annotation. Red shaded regions denote intervals selected by the annotation agent, the blue area shows the timestep-wise action MSE of an example policy, and the sampled frames illustrate the annotated interval boundaries. Left: simulation task \emph{place\_apple\_from\_bowl\_into\_bin}; right: real-world task \emph{pick\_place\_mouse}.}
\label{fig:annotation_examples}
\end{figure}

We automate the localization of critical intervals with a zero-shot agent skill. The VLM agent first scans the complete local video timeline, then inspects denser frames around candidate boundaries. A fixed episode protocol defines three general criteria: purposeful object interaction, precision near a target, and visibly high-difficulty or high-risk manipulation. Fig.~\ref{fig:annotation_examples} shows example annotations in simulation and the real world; Appendix~\ref{subsec:agent_skill} shows the annotation protocol.

\subsection{Action Alignment for Offline Validation}
\label{subsec:alignment}

Offline validation should reflect not only \emph{what} actions a policy predicts, but also \emph{how} those predictions are executed at test time. We therefore incorporate simple action-alignment procedures that better match rollout settings.

\paragraph{Temporal Ensembling and RTC.} Inference-time methods such as temporal ensembling or real-time action chunking (RTC) are used during rollout to produce smooth execution. These methods can materially alter policy behavior. We therefore apply the same deployment-time inference methods to offline validation to better match rollout-time behavior and partially model the effect of compounding errors. To apply these procedures in offline validation, we validate over monotonic timesteps within a trajectory and apply temporal ensembling or RTC using overlapping action chunks from previous predictions. Temporal ensembling smooths rollout trajectories by averaging action predictions from overlapping chunks. 
\begin{equation}
    f(o_t)^{\mathrm{TE}}_i
    = \frac{1}{H} \sum_{h=0}^{H-1} f(o_{t-h})_{i+h}
\end{equation}
where $f(o_t)_i$ is the action for step $i$ from the chunk predicted with observation $o_t$, and $H$ is the ensemble horizon. Under high control frequency, temporal ensembling can also be interpreted as repeated sampling, which reduces variance in stochastic predictions. RTC guides the action generation process with overlapping action chunks.
Let $\Delta$ be the number of delayed controller steps caused by inference latency during rollout. Applying RTC to offline validation can be viewed as inpainting $f(o_t)$ while softly matching the overlapping actions from $f(o_{t-\Delta})^{\mathrm{RTC}}$. After aligning the predictions using rollout-time TE or RTC, we compute the validation error against expert actions on the aligned action chunks.

\paragraph{Dynamic Time Warping (DTW).} We use dynamic time warping (DTW)~\cite{sakoe1978dynamic} to compare predicted and expert action sequences under small temporal misalignments. Rather than matching each predicted action only to the expert action at the same timestep, DTW finds a monotone alignment path $P$ that minimizes accumulated action error. For each critical timestep, we define the timestep-level CI-MSE as:
\begin{equation}
\ell_{\mathrm{CI}}(f,o_t, a_t) =
\min_{P \in \mathcal{P}_w}
\frac{1}{|P|}
\sum_{(p,q)\in P} \| f(o_t)_p^{\mathrm{TE/RTC}} - a_{t,q} \|_2^2,
\end{equation}
where $\mathcal{P}_w$ denotes the set of monotone warping paths satisfying $|p-q|\leq w$ for window size $w$. This is useful because manipulation demonstrations often vary in tempo: two executions can make the same contact, grasp, or release slightly earlier or later while remaining behaviorally equivalent. A pointwise MSE would penalize such phase shifts even when the underlying action sequence is correct. We therefore apply DTW only as a local alignment correction, with $w$ treated as a validation hyperparameter, so that the metric discounts harmless timing offsets without allowing large reorderings or failures to be hidden by excessive warping.

The dataset-level CI-MSE is then obtained by aggregating this timestep-level error over the critical validation set:
\begin{equation}
L_{\mathrm{CI}}(f;\mathcal{D}_{\mathrm{val}})
=
\mathcal{A}
\left(
\left\{
\ell_{\mathrm{CI}}(f, o_{i,t}, a_{i,t})
:
(o_{i,t}, a_{i,t}) \in \mathcal{D}_{\mathrm{crit}}
\right\}
\right).
\end{equation}
For each selected, DTW-aligned frame, the squared $\ell_2$ error aggregates all action dimensions without dimension-wise normalization. Here $\mathcal{A}$ is the median over the resulting frame-level errors across validation trajectories; raw MSE uses the same aggregation over the full trajectories, without critical-interval filtering or alignment unless an ablation explicitly states otherwise.

\subsection{Ordinal Rollout Evaluation with Elo}
\label{subsec:elo}

This module is not part of CI-MSE; rather, it enhances the evaluation protocol. In real-world experiments, researchers often use partial-success or progress scores when binary success rates are too coarse and the number of available trials is small. Such scalar scores implicitly assume meaningful numerical gaps between adjacent progress levels, which may be unjustified. We instead use an Elo-style ranking system~\cite{elo1978rating} in which only the ordinal ordering of two trial outcomes determines a win or tie. Although Elo has no explicit stability regularizer, our experiments demonstrate its empirical stability, as shown in Appendix~\ref{subsec:rollout_scores}.

%% file: sections/experiments.tex
\section{Experiments}
\label{sec:experiments}

\subsection{Experimental Setup}
\label{subsec:setup}

We evaluate CI-MSE in both simulation and real-world settings. In simulation, we use the LBM-Eval benchmark and its associated datasets~\cite{barreiros2026careful}, which consist of 49 tasks and approximately 10,000 demonstrations. We compare controlled vision-language-action policy variants that differ in architecture, dataset scale, training steps, parameter-efficient fine-tuning, action-head size, and VLM backbone, as summarized in Table~\ref{tab:sim_variant_config}. In the real world, we study diffusion policies trained on data-scaling datasets~\cite{lin2025data} and evaluated on a Franka arm across the pour-water, arrange-mouse, fold-towel, and unplug tasks. Table~\ref{tab:cimse_hyperparams} lists the CI-MSE hyperparameters for both settings. The evaluated chunk steps differ because the settings use different controller frequencies and predicted action-chunk lengths. The compositions of the training and validation sets are described in Appendix~\ref{subsec:data_composition}.

For simulation, rollout performance is measured by success rate. For real-world experiments, where policy gaps are smaller and trial budgets are tighter, we use Elo scores as the main rollout score and compare them against partial success scores; Appendix~\ref{subsec:rollout_scores} analyzes the stability of these rollout scores. Offline validation compares raw MSE with CI-MSE as defined in Section~\ref{sec:method}. In both simulation and real-world experiments, we additionally report Action Magnitude-Normalized MSE (AMN-MSE) as a baseline that controls for differences in action scale.

To measure the reliability of the validation metrics, we report correlation and top-1 selection regret. The correlation coefficients---Pearson's $r$ and Spearman's $\rho$---reflect the global agreement between validation error and rollout score. The top-1 selection regret $R$ is the difference in rollout score between the best rollout-performing policy and the policy selected by the validation metric; it reflects the metric's ability to select the optimal policy.

\begin{table}[t]
\centering
\caption{CI-MSE hyperparameters. Evaluated chunk steps denote the indices in a predicted chunk used for execution or error computation.}
\label{tab:cimse_hyperparams}
\small
\setlength{\tabcolsep}{4pt}
\begin{tabular}{lcc}
\toprule
Hyperparameters & Simulation & Real-world \\
\midrule
Ensemble horizon & 8 & 8 \\
DTW window size & 1 & 1 \\
Evaluated chunk steps & 0--24 & 2--8 \\
\bottomrule
\end{tabular}
\end{table}

\begin{figure}[t]
\centering
\includegraphics[width=\linewidth]{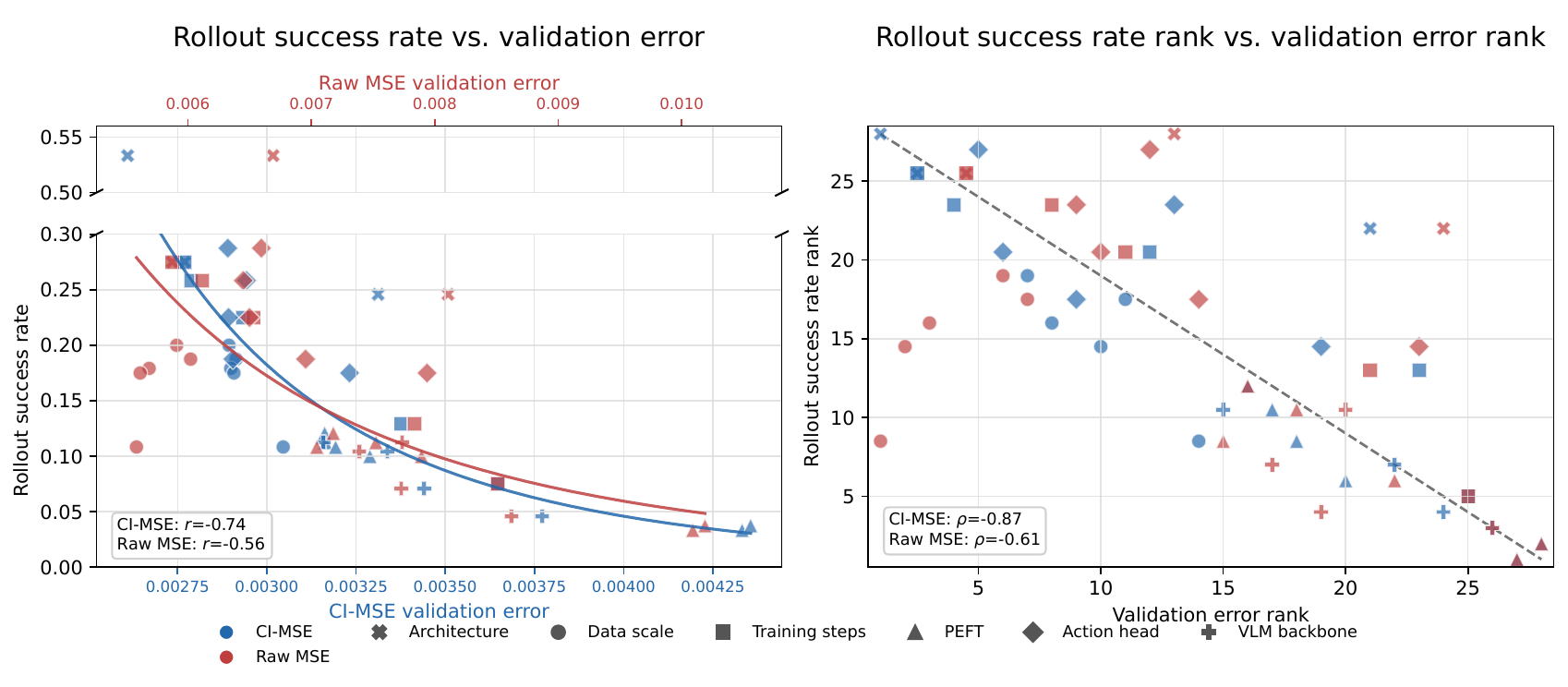}
\caption{Simulation evaluation-validation correlation. Left: evaluation success rate versus validation error, with CI-MSE and raw MSE shown on aligned validation-error scales. The fitting curve follows the power law $y = ax^b$. Right: rank correlation between success-rate rank and validation-error rank. Marker color indicates the validation metric, and marker shape indicates the model-variant group.}
\label{fig:main_corr}
\end{figure}

\subsection{Correlation between Validation Error and Rollout Performance in Simulation}
\label{subsec:sim_corr}

\begin{table}[t]
\centering
\caption{Simulation model variants. All non-architecture variants are based on X-VLA.}
\label{tab:sim_variant_config}
\small
\setlength{\tabcolsep}{3pt}
\begin{tabular}{llc}
\toprule
Variant family & Values & Number of checkpoints \\
\midrule
Architecture & $\pi_{0.5}$\cite{black2025pi_}, X-VLA\cite{zheng2025x}, Gr00t N1.7\cite{bjorck2025gr00t} & 3 \\
Data scale & 20\%, 40\%, 60\%, 80\%, 100\% of training data & 5 \\
Training steps & 20k, 40k, 60k, 80k, 100k & 5 \\
PEFT & LoRA on \{all, QKV, FFN\} layers; rank \{8, 16\} & 6 \\
Action head & 175M, 310M, 580M, 930M, 1220M parameters & 5 \\
VLM backbone & florence2-\{base, large\}, paligemma2-3B-\{pt, mix\} & 4 \\
\bottomrule
\end{tabular}
\end{table}

Our first experiment measures how well offline metrics predict rollout outcomes across VLA model variants. Overall, Fig.~\ref{fig:main_corr} shows a substantial improvement in validation--rollout agreement. Across 27 distinct policies, CI-MSE achieves $r=-0.74$ and $\rho=-0.87$, whereas raw MSE reaches $r=-0.56$ and $\rho=-0.61$. Table~\ref{tab:sim_variants} gives the corresponding correlations by variant family, showing that agreement is not uniform across variant types. For example, when training steps vary, all three metrics are nearly perfectly rank-consistent with rollout success. However, when data scale varies, raw MSE and AMN-MSE are inverted and have positive rank correlations ($\rho=0.90$), even though lower validation error should predict higher rollout success. CI-MSE instead achieves $r=-0.97$ and $\rho=-0.70$ for this family. In addition, CI-MSE achieves zero regret@1 for all variant families, indicating that it selects the optimal policy along every model axis. AMN-MSE achieves slightly stronger correlations than raw MSE, but its regret@1 remains high and its overall performance remains well below that of CI-MSE. This result indicates that the benefit of critical-interval filtering cannot be explained merely by the small magnitude of actions in those intervals.

This suggests that uniformly averaged MSE remains unreliable for certain variant families even after normalizing for action magnitude. When model quality changes through data scale, backbone choice, or other factors that alter policy behavior, averaging error over all timesteps can let irrelevant motion obscure the success signal. CI-MSE alleviates this issue across the examined variant families.

\begin{table}[t]
\centering
\caption{Correlation and regret across simulation variant families. Blue cells indicate negative correlation, and orange cells indicate positive correlation. Color intensity indicates correlation magnitude. Best values are highlighted in bold.}
\label{tab:sim_variants}
\small
\setlength{\tabcolsep}{6pt}
\begin{tabular}{llccc}
\toprule
Variant family & Validation metric & $r\downarrow$ & $\rho\downarrow$ & $R\downarrow$ \\
\midrule
\multirow{3}{*}{Architecture} & Raw MSE & \corrneg{17}{-0.24} & \corrneg{35}{-0.50} & 25.83 \\
 & AMN-MSE & \corrneg{42}{-0.59} & \corrneg{70}{-1.00} & \best{0.00} \\
 & \textbf{CI-MSE} & \corrneg{52}{\best{-0.74}} & \corrneg{70}{\best{-1.00}} & \best{0.00} \\
\midrule
\multirow{3}{*}{Data scale} & Raw MSE & \corrpos{47}{0.67} & \corrpos{63}{0.90} & 9.17 \\
 & AMN-MSE & \corrpos{48}{0.68} & \corrpos{63}{0.90} & 9.17 \\
 & \textbf{CI-MSE} & \corrneg{68}{\best{-0.97}} & \corrneg{49}{\best{-0.70}} & \best{0.00} \\
\midrule
\multirow{3}{*}{Training steps} & Raw MSE & \corrneg{70}{-1.00} & \corrneg{70}{-1.00} & 0.00 \\
 & AMN-MSE & \corrneg{70}{-1.00} & \corrneg{70}{-1.00} & 0.00 \\
 & \textbf{CI-MSE} & \corrneg{70}{-1.00} & \corrneg{70}{-1.00} & 0.00 \\
\midrule
\multirow{3}{*}{PEFT} & Raw MSE & \corrneg{69}{-0.99} & \corrneg{54}{-0.77} & 1.25 \\
 & AMN-MSE & \corrneg{69}{-0.99} & \corrneg{54}{-0.77} & 1.25 \\
 & \textbf{CI-MSE} & \corrneg{70}{\best{-0.99}} & \corrneg{66}{\best{-0.94}} & \best{0.00} \\
\midrule
\multirow{3}{*}{Action head} & Raw MSE & \corrneg{53}{-0.75} & \corrneg{49}{-0.70} & 2.92 \\
 & AMN-MSE & \corrneg{53}{\best{-0.75}} & \corrneg{49}{-0.70} & 2.92 \\
 & \textbf{CI-MSE} & \corrneg{42}{-0.60} & \corrneg{49}{-0.70} & \best{0.00} \\
\midrule
\multirow{3}{*}{VLM backbone} & Raw MSE & \corrneg{58}{-0.83} & \corrneg{28}{-0.40} & 0.83 \\
 & AMN-MSE & \corrneg{59}{-0.84} & \corrneg{56}{-0.80} & 0.83 \\
 & \textbf{CI-MSE} & \corrneg{67}{\best{-0.96}} & \corrneg{70}{\best{-1.00}} & \best{0.00} \\
\bottomrule
\end{tabular}
\end{table}

\subsection{Distribution Shift in Policy Evaluation}
\label{subsec:ood}

We next study a harder regime in which validation and evaluation are matched but differ from the training distribution. This setting captures a common use case: researchers want to validate out-of-distribution (OOD) generalization. Table~\ref{tab:ood} is structured around three distribution-shift dimensions: object-layout OOD, visual OOD, and skill OOD. For object-layout OOD, objects are spawned outside the bounding box used in the training dataset. For visual OOD, we change the background and table texture. For skill OOD, we select five tasks unseen in the training set.

Table~\ref{tab:ood} shows that CI-MSE remains the strongest global rank predictor under object-layout and skill shifts. The advantage is largest for skill shift, where CI-MSE improves rank correlation from $\rho=-0.36$ to $\rho=-0.69$. The correlations for skill and visual OOD are relatively low because many model variants achieve near-zero success rates, making these variants hard to separate. All three metrics obtain zero regret@1 under object-layout shift, and CI-MSE still improves Spearman correlation from $\rho=-0.77$ to $\rho=-0.88$. Under visual shift, CI-MSE has the lowest regret at 1.39 percentage points. Under skill shift, AMN-MSE selects the rollout-optimal policy despite having weaker global rank correlation than CI-MSE, again illustrating that correlation and regret@1 capture different behavior. Researchers should use these offline metrics cautiously when evaluating visual and skill-level generalization.

\begin{table}[t]
\centering
\caption{Correlation and regret under evaluation distribution shift on simulation policies.}
\label{tab:ood}
\small
\setlength{\tabcolsep}{4pt}
\begin{tabular}{l*{9}{c}}
\toprule
& \multicolumn{9}{c}{Distribution shift types} \\
\cmidrule(lr){2-10}
\makecell[l]{Validation\\metric}
& \multicolumn{3}{c}{\makecell{Object\\layout}}
& \multicolumn{3}{c}{Visual}
& \multicolumn{3}{c}{Skill} \\
\cmidrule(lr){2-4}
\cmidrule(lr){5-7}
\cmidrule(lr){8-10}
& $r\downarrow$ & $\rho\downarrow$ & $R\downarrow$
& $r\downarrow$ & $\rho\downarrow$ & $R\downarrow$
& $r\downarrow$ & $\rho\downarrow$ & $R\downarrow$ \\
\midrule
Raw MSE
& \corrneg{54}{-0.77} & \corrneg{54}{-0.77} & 0.00
& \corrneg{41}{-0.58} & \corrneg{48}{\best{-0.68}} & 12.50
& \corrneg{20}{-0.29} & \corrneg{25}{-0.36} & 12.00 \\
AMN-MSE
& \corrneg{53}{-0.76} & \corrneg{60}{-0.86} & 0.00
& \corrneg{38}{-0.54} & \corrneg{44}{-0.63} & 32.50
& \corrneg{25}{-0.36} & \corrneg{47}{-0.67} & \best{0.00} \\
\textbf{CI-MSE}
& \corrneg{57}{\best{-0.81}} & \corrneg{62}{\best{-0.88}} & 0.00
& \corrneg{43}{\best{-0.62}} & \corrneg{46}{-0.66} & \best{1.39}
& \corrneg{34}{\best{-0.48}} & \corrneg{48}{\best{-0.69}} & 12.00 \\
\bottomrule
\end{tabular}
\end{table}

\subsection{Validation-Rollout Correlation and Cross-Domain Effects in Real-World Experiments}
\label{subsec:real_corr}

The real-world experiments test whether the simulation trend carries over to real physical evaluation and limited trials. Evaluated diffusion policies are trained on $2^m$ randomly selected object-environment pairs $(m=2,3,4,5)$~\cite{lin2025data}. Table~\ref{tab:real_corr} reports the corresponding correlations and regret@1 in Elo points. CI-MSE is the stronger predictor for most cases: it achieves near-perfect cross-environment agreement for pour water and arrange mouse, and it substantially improves fold-towel cross-object validation over raw MSE and AMN-MSE. AMN-MSE selects the same top-ranked policy as raw MSE in all eight task-shift settings. CI-MSE reduces regret@1 from 222.92 to 0 Elo points for pour-water cross-environment evaluation and from 28.90 to 0 for fold-towel cross-object evaluation.

We also investigate the effect of data collectors' action styles on validation error. The fold-towel and unplug settings introduce an additional practical confounder: because the original dataset does not include validation sets for these two tasks, their validation datasets were collected by operators who were different from those who collected the training datasets. This makes the validation error harder to interpret: a policy can receive a higher offline error because it does not match the validation operator's style, even if its rollout behavior remains competitive under the evaluation protocol. Under this confounder, correlations are weaker and less consistent. This indicates that consistent data-collection protocols are important when using action-space validation error as a proxy for real-world performance.

\begin{table*}[t]
\centering
\caption{Correlation and regret for real-world rollout Elo score. Each task is evaluated across unseen environments (Env.) and unseen objects (Obj.).}
\label{tab:real_corr}
\small
\setlength{\tabcolsep}{2pt}
\resizebox{\textwidth}{!}{%
\begin{tabular}{l*{24}{c}}
\toprule
& \multicolumn{12}{c}{Collectors matched}
& \multicolumn{12}{c}{Collectors mismatched} \\
\cmidrule(lr){2-13}
\cmidrule(lr){14-25}
& \multicolumn{6}{c}{Pour water}
& \multicolumn{6}{c}{Arrange mouse}
& \multicolumn{6}{c}{Fold towel}
& \multicolumn{6}{c}{Unplug} \\
\cmidrule(lr){2-7}
\cmidrule(lr){8-13}
\cmidrule(lr){14-19}
\cmidrule(lr){20-25}
\makecell[l]{Validation\\metric}
& \multicolumn{3}{c}{Env.}
& \multicolumn{3}{c}{Obj.}
& \multicolumn{3}{c}{Env.}
& \multicolumn{3}{c}{Obj.}
& \multicolumn{3}{c}{Env.}
& \multicolumn{3}{c}{Obj.}
& \multicolumn{3}{c}{Env.}
& \multicolumn{3}{c}{Obj.} \\
\cmidrule(lr){2-4}
\cmidrule(lr){5-7}
\cmidrule(lr){8-10}
\cmidrule(lr){11-13}
\cmidrule(lr){14-16}
\cmidrule(lr){17-19}
\cmidrule(lr){20-22}
\cmidrule(lr){23-25}
& $r\downarrow$ & $\rho\downarrow$ & $R\downarrow$
& $r\downarrow$ & $\rho\downarrow$ & $R\downarrow$
& $r\downarrow$ & $\rho\downarrow$ & $R\downarrow$
& $r\downarrow$ & $\rho\downarrow$ & $R\downarrow$
& $r\downarrow$ & $\rho\downarrow$ & $R\downarrow$
& $r\downarrow$ & $\rho\downarrow$ & $R\downarrow$
& $r\downarrow$ & $\rho\downarrow$ & $R\downarrow$
& $r\downarrow$ & $\rho\downarrow$ & $R\downarrow$ \\
\midrule
Raw MSE
& \corrneg{33}{-0.47} & \corrneg{56}{-0.80} & 222.92
& \corrneg{60}{-0.86} & \corrneg{70}{-1.00} & 0.00
& \corrneg{56}{-0.80} & \corrneg{56}{-0.80} & 0.00
& \corrneg{5}{-0.06} & \corrneg{14}{-0.20} & 180.21
& \corrpos{19}{0.27} & \corrpos{42}{0.60} & 135.75
& \corrneg{6}{-0.09} & \corrpos{5}{0.00} & 28.90
& \corrpos{43}{0.62} & \corrpos{28}{0.40} & 357.90
& \corrneg{60}{\best{-0.86}} & \corrneg{56}{\best{-0.80}} & 23.48 \\
AMN-MSE
& \corrneg{32}{-0.46} & \corrneg{56}{-0.80} & 222.92
& \corrneg{61}{-0.87} & \corrneg{70}{\best{-1.00}} & 0.00
& \corrneg{55}{-0.79} & \corrneg{56}{-0.80} & 0.00
& \corrneg{6}{-0.08} & \corrneg{14}{-0.20} & 180.21
& \corrpos{19}{0.28} & \corrpos{42}{0.60} & 135.75
& \corrneg{7}{-0.10} & \corrpos{5}{0.00} & 28.90
& \corrpos{43}{0.62} & \corrpos{28}{0.40} & 357.90
& \corrneg{59}{-0.85} & \corrneg{56}{\best{-0.80}} & 23.48 \\
\textbf{CI-MSE}
& \corrneg{69}{\best{-0.99}} & \corrneg{70}{\best{-1.00}} & \best{0.00}
& \corrneg{70}{\best{-0.99}} & \corrneg{70}{\best{-1.00}} & 0.00
& \corrneg{67}{\best{-0.96}} & \corrneg{70}{\best{-1.00}} & 0.00
& \corrneg{46}{\best{-0.66}} & \corrneg{28}{\best{-0.40}} & 180.21
& \corrneg{5}{\best{-0.05}} & \corrneg{28}{\best{-0.40}} & 135.75
& \corrneg{61}{\best{-0.87}} & \corrneg{70}{\best{-1.00}} & \best{0.00}
& \corrpos{34}{\best{0.48}} & \corrpos{28}{0.40} & 357.90
& \corrneg{37}{-0.53} & \corrneg{42}{-0.60} & 23.48 \\
\bottomrule
\end{tabular}
}
\end{table*}

\subsection{Ablation Study and Sensitivity Analysis}
\label{subsec:sensitivity}

To isolate each component's contribution, we add components to raw MSE and remove them from CI-MSE, as shown in Table~\ref{tab:component_ablation}. Critical intervals provide the largest individual rank-correlation gain ($-0.61\rightarrow-0.78$), as well as the largest reduction in regret@1 (42.50 to 25.80 percentage points). TE is complementary, whereas DTW gives a smaller improvement in correlation. Removing TE or DTW from CI-MSE retains zero regret, while removing critical intervals raises regret to 35.38\%. 

\begin{table}[t]
\centering
\caption{Component ablations. $\Delta\rho$ is the gap in Spearman's $\rho$ to full CI-MSE.}
\label{tab:component_ablation}
\small
\setlength{\tabcolsep}{3pt}
\begin{tabular}{lcccc}
\toprule
Variant & $r\downarrow$ & $\rho\downarrow$ & $\Delta\rho\downarrow$ & $R\downarrow$ \\
\midrule
Raw MSE & -0.56 & -0.61 & +0.26 & 42.50 \\
Raw MSE $+$ Critical Interval & -0.66 & -0.78 & +0.09 & 25.80 \\
Raw MSE $+$ TE & -0.63 & -0.67 & +0.20 & 35.80 \\
Raw MSE $+$ DTW & -0.62 & -0.68 & +0.19 & 42.47 \\
\addlinespace[1pt]
CI-MSE $-$ Critical Interval & -0.59 & -0.69 & +0.18 & 35.38 \\
CI-MSE $-$ TE & -0.67 & -0.80 & +0.07 & \best{0.00} \\
CI-MSE $-$ DTW & -0.74 & -0.85 & +0.02 & \best{0.00} \\
\midrule
\textbf{CI-MSE full} & \textbf{-0.74} & \textbf{-0.87} & -- & \textbf{0.00} \\
\bottomrule
\end{tabular}
\end{table}

\begin{figure}[t]
\centering
\includegraphics[width=\linewidth]{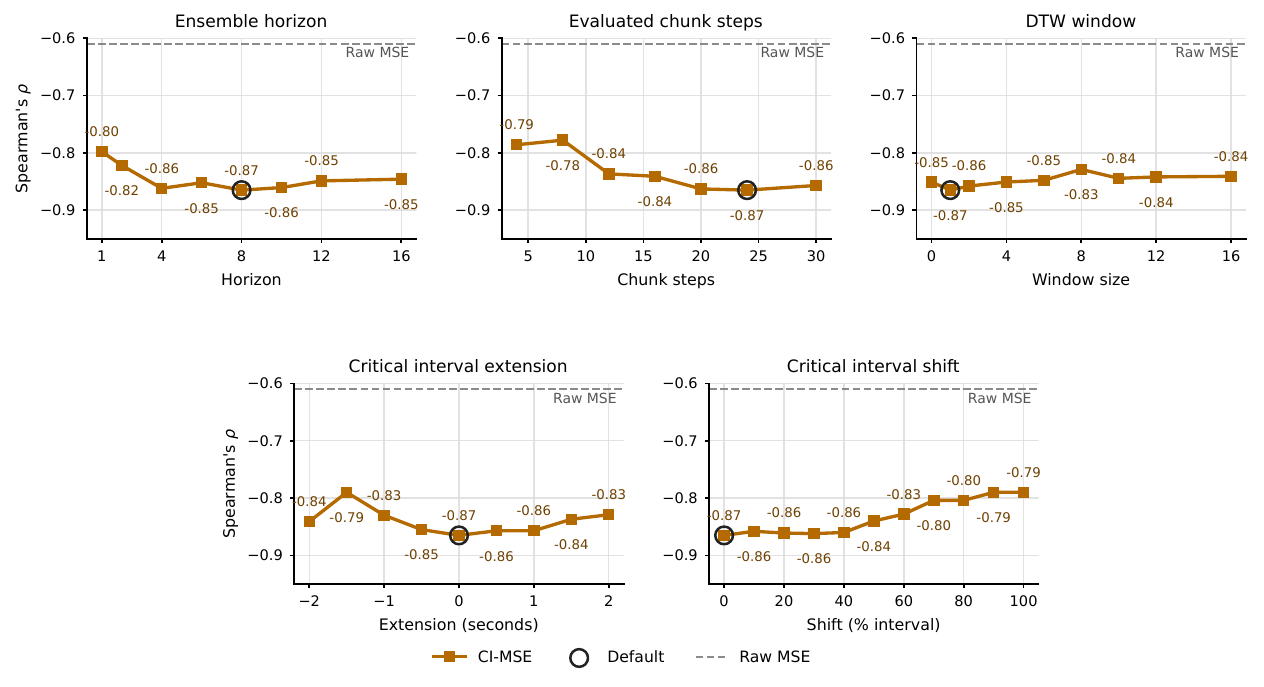}
\caption{Sensitivity analysis of CI-MSE to its hyperparameters and interval annotation.}
\label{fig:sens}
\end{figure}

We perform a sensitivity analysis of CI-MSE to its hyperparameters and critical interval annotation. We vary the ensemble horizon, evaluated action chunk steps, DTW window size, and critical interval length, and we shift the critical interval. This sweep also includes component ablations: setting the ensemble horizon to $H=1$ removes temporal ensembling, and setting the DTW window size to $W=0$ removes DTW. We then compute the correlation and regret between CI-MSE and rollout success rate using the same model checkpoints and validation set as Section~\ref{subsec:sim_corr}. Fig.~\ref{fig:sens} shows that CI-MSE is relatively insensitive to these choices across a reasonable range of hyperparameters and consistently outperforms raw MSE.

The sensitivity trends are consistent with the design of CI-MSE. Temporal ensembling improves correlation from the no-ensembling ablation to $H=8$, after which the curve remains strong but slightly weaker. DTW window size $W=1$ gives the best result compared with the no-warping ablation at $W=0$, while larger windows gradually weaken the correlation. For interval length, slightly extending or shortening the critical interval (<1 second) barely degrades the correlation. For the shift experiment, the horizontal value is the temporal displacement divided by the annotated interval length; we randomly shift intervals forward or backward. Correlation remains stable below a 40\% displacement and degrades for larger shifts. Even a 100\% shift can retain weaker outcome information because temporally adjacent approach or recovery actions may still reflect policy quality, but it remains worse than using the annotated interval. This indicates that the critical interval is the core of our metric and that it tolerates moderate annotation imprecision. Overall, the metric is not brittle in the useful operating range, indicating that the improvement is not driven by a narrowly tuned hyperparameter choice.

%% file: sections/conclusion.tex
\section{Conclusion and Limitations}
\label{sec:conclusion}
\label{sec:limitations}

This paper introduced \emph{critical interval MSE}, an offline validation metric that filters task-irrelevant action errors and matches rollout-time behavior. Across simulation and real-world experiments, CI-MSE correlates better with rollout outcomes than raw MSE, including under several distribution shifts, and remains stable across reasonable hyperparameter choices.

CI-MSE still inherits the limits of offline validation: it does not observe dynamics, so it can misread policies that solve a task through valid approaches not represented in the demonstrations. It assumes a reasonably consistent action mode between training and validation; operator or collection-protocol mismatch can affect validation quality. In addition, our current focus is on short-horizon manipulation; CI-MSE is less suitable for long-horizon tasks that depend on planning or memory. Despite these limitations, our method is applicable to a wide range of manipulation tasks and offers a practical tool for comparing nearby model variants and accelerating policy iteration.

%% file: sections/acknowledgments.tex
\section*{Acknowledgments}

This research was conducted with the support of the Shanghai Qi Zhi Institute \& Spirit AI Innovation Program and the Tsinghua University Dushi Program. Funding and support for this work were also provided by the Tsinghua University - Keystone Electrical (Zhejiang) Co.,Ltd Joint Research Center for Embodied Multimodal Artificial Intelligence (JCEMAI). Additionally, we would like to extend our thanks to the Xiongan AI Institute. We thank Yingdong Hu for helpful guidance on writing and presentation. We thank Fanqi Lin for valuable guidance on the experimental setup.

%% file: sections/appendix.tex
\section{Appendix}
\label{sec:appendix}

\subsection{Zero-Shot Agent Skill for Critical-Interval Annotation}
\label{subsec:agent_skill}

We annotate critical intervals with the self-contained agent skill reproduced below. The skill uses no task-specific annotation examples and no external annotation API. We reproduce its orchestration instructions and complete per-episode annotation protocol below. Paths mentioned in the listings refer to the annotation skill package.

\subsubsection*{Agent skill.}
\begin{lstlisting}[style=skilldoc]
---
name: critical-interval-annotator
description: Annotate outcome-critical time intervals in one or more robot manipulation video episodes using local visual inspection. Use for zero-shot critical-interval labeling from video files or dataset episodes; do not use for metadata-only labeling or API-based VLM annotation.
---

# Critical Interval Annotator

Orchestrate local, zero-shot annotation while keeping video content out of the main agent's context.

## Non-negotiable isolation

- Treat each video file or requested dataset episode as one episode.
- Spawn a **fresh subagent for every episode**. Never assign two episodes to one agent and never reuse an episode agent for a later episode.
- The main agent must not open, play, sample, transcode, screenshot, extract frames from, or visually inspect any video. It may inspect filenames, paths, episode indices, and non-visual dataset metadata solely to enumerate work, and it may read the final structured labels and timestamps.
- Only the assigned episode agent may inspect that episode and its derived frames. Pass it the episode source, identifier, optional task instruction, output path, and [episode protocol](references/episode-protocol.md).
- Do not upload video or call Gemini, Vertex AI, OpenAI API, Hugging Face downloads, web services, or any other external annotation API. Dataset sources must already exist locally. Local commands and local image inspection are allowed inside the episode agent.
- Keep the skill self-contained: its scripts must not import `vlm_annotator` or other modules from the host repository. They may use installed local runtime libraries such as `zarr`, NumPy, OpenCV, and `ffmpeg`.
- If subagent delegation is unavailable, stop and explain that the isolation requirement cannot be satisfied; do not inspect the video in the main thread.

## Orchestration

1. Resolve the episodes without examining their visual content. For direct inputs, each video is an episode. For a dataset, use the user's requested episode range; if none is given, use only an unambiguous single episode or ask for the range when processing all episodes would be unexpectedly large.
2. Create a new, run-specific result directory outside the source data. Assign every episode a unique non-negative integer identifier, preserving dataset indices when available. Do not reuse a result directory or modify source videos or datasets.
3. Dispatch as many fresh episode agents concurrently as available slots permit, then continue in waves. Each task must tell the agent to:
   - read `references/episode-protocol.md` in this skill;
   - inspect only its assigned episode using local tools;
   - incorporate the user's optional task instruction as context;
   - write its result atomically to `<result-dir>/episode-<episode-id>.json` using the protocol schema;
   - return a short status without embedding contact sheets or video frames in its final response.
4. Wait for every episode agent. A user-input request from one agent concerns only that episode; do not let another agent inspect it as a shortcut. A retry after an agent failure must use a fresh replacement agent assigned only to that same episode.
5. The main agent may read the final structured JSON labels and timestamps. It must not read intermediate images, contact sheets, extracted clips, or free-form visual inspection notes.
6. Validate and merge results with:

   ```text
   python scripts/validate_and_merge.py --input-dir <result-dir> --expected-episodes <id> [<id> ...] --output <result-dir>/annotations.json --seconds-output <result-dir>/annotations_seconds.json
   ```

   Resolve script paths relative to this skill directory. Validation failure should go back to a fresh replacement episode agent; the main agent must not infer or repair timestamps from video.
   This canonicalizes each per-episode JSON, writes `annotations_seconds.json` with auditable second-based boundaries, and writes pipeline-compatible `annotations.json` with boundaries converted to frame indices.
7. Return both merged JSON paths, episode count, and any failed episode identifiers. For a single episode, the canonicalized per-episode JSON is also a direct seconds-based annotation artifact.

Keep an optional task instruction verbatim when handing off. It helps infer the intended objects and target state but does not make the entire task critical.
\end{lstlisting}
\subsubsection*{Episode annotation protocol.}
\begin{lstlisting}[style=skilldoc]
# Episode annotation protocol

This protocol is for the fresh subagent assigned to exactly one episode. Do not inspect any other episode and do not call an external API or upload the video.

## Inputs

The orchestration task provides:

- one video path, or one dataset path plus episode index;
- a unique non-negative integer episode identifier;
- an output JSON path;
- an optional natural-language task instruction.

For a local zarr or LeRobot v2.1 dataset, materialize only the assigned episode with `scripts/materialize_episode.py`. Pass the dataset type, existing local dataset path, episode index, temporary output path, and desired downsample factor. The self-contained helper reads zarr arrays or LeRobot metadata/video files directly; it does not import repository modules or use a dataset hub. It reports original FPS, frame count, and duration; retain the original FPS and frame count for output conversion. Never edit the source dataset.

## Inspect locally

Use local tools such as `ffprobe` and `ffmpeg` to determine duration and extract frames or short clips into a temporary directory. Inspect those derived images yourself with the local image-viewing tool.

Treat source time as authoritative: frame `i` is at `i / fps` seconds. Downsampling does not change this mapping. For a downsampled inspection MP4, ignore any encoded tail beyond `total_frames / fps`; it is an artifact of the final sampled frame.

First inspect the full timeline at a coarse cadence sufficient to find contacts, target approaches, retries, and distinct subgoals. Then inspect denser frames around every candidate boundary; use approximately 0.1--0.25 second spacing when the source supports it. Scan the entire episode before applying the retry rule. Do not label from filenames, metadata, task text, or a single still image alone.

The main agent is prohibited from seeing the video, frames, contact sheets, or extracted clips. Keep all visual inspection and temporary visual artifacts inside this episode task. Store temporary media outside the result directory and remove it when practical; never remove source data.

## Critical interval definition

Identify the shortest intervals that most directly determine whether the manipulation succeeds or fails.

An interval is critical when it contains at least one of these:

1. **Object interaction.** The robot makes an imminent purposeful grasp, push, pull, or other contact. Start approximately 5 cm before contact, or at the final deliberate approach when distance is not visually reliable. End when the intended contact is established or completed.
2. **Precision near a target.** The robot performs fine positioning, orientation, placement, insertion, or alignment near a required target. Start approximately 10 cm from the target, or at the visible transition from coarse transport to careful alignment. End at a stable intended pose, achieved insertion/alignment, or clear failure.
3. **High difficulty or risk.** The robot manipulates a fragile, deformable, unstable, or easily dropped object; traverses a narrow constraint; maintains delicate contact; overcomes resistance; or avoids a close obstacle. Include only the visibly risky portion.

Do not annotate long unconstrained transit, idle stabilization after success, homing/reset, exploratory or random positioning unrelated to the goal, ordinary non-contact motion, or post-action retreat unless retreat is required for success.

## Boundaries and retries

- One interval represents one coherent critical action.
- Merge adjacent or overlapping phases of one continuous action, such as final approach plus grasp closure.
- Separate distinct actions when non-critical motion occurs between them.
- Exclude unnecessary lead-in and follow-through. Intervals must not overlap and must be chronological.
- For repeated attempts at the same immediate subgoal, keep only the final attempt. A new attempt begins after disengagement, backing away, reset, or a clearly renewed approach.
- For multiple distinct required subgoals, keep the final attempt for each subgoal.
- If success is unclear, use the last visible attempt.
- If no critical interval is visible, return an empty list.

Use the optional task instruction only to infer the intended goal, objects, and target state. Apply the criteria independently; do not label the full instructed task by default.

## Output

Estimate boundaries from visible motion, contact state, and object motion. Round to the nearest 0.1 seconds and ensure `0.0 <= start < end <= duration_seconds`.

Write one UTF-8 JSON object atomically to the assigned output path:

```json
{
  "episode": 0,
  "source": "<provided source path or dataset-path#episode-index>",
  "duration_seconds": 12.3,
  "fps": 30.0,
  "total_frames": 369,
  "intervals": [
    {
      "label": "secure grasp of the cup",
      "start": 2.3,
      "end": 4.7
    }
  ]
}
```

Use only these top-level and interval fields. `episode` must be the provided integer identifier. `fps` and `total_frames` describe the original episode, not the downsampled inspection encoding. Use concise, action-specific labels. JSON numbers do not preserve display precision, but write one decimal digit in the serialized text when possible. Do not include reasoning, confidence, frame paths, free-form inspection notes, or task instructions in the JSON; the concise action label is permitted and required.

Before finishing, parse the file locally and verify its schema, finite timestamps, duration bounds, chronological order, and lack of overlap. Return only a short completion status and the JSON path to the main agent.
\end{lstlisting}

\subsection{Revisiting Rollout Scores}
\label{subsec:rollout_scores}

\begin{figure}[H]
\centering
\includegraphics[width=\linewidth]{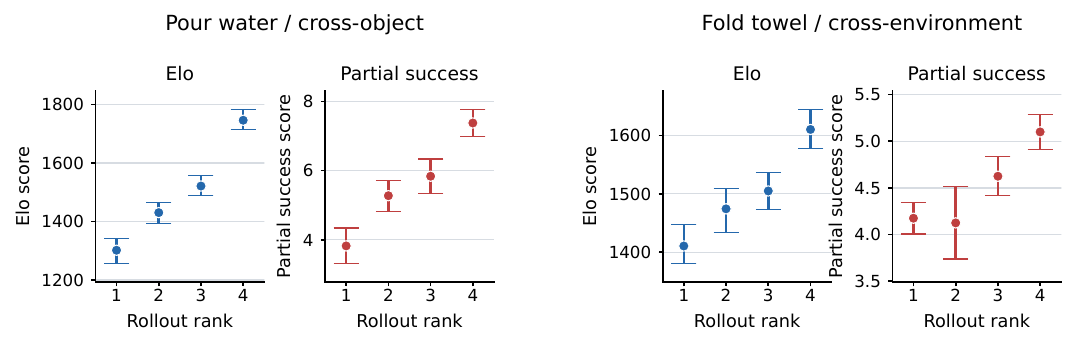}
\caption{Confidence intervals for Elo scores and partial success scores. The confidence intervals are computed via 1,000 bootstrap resamples. Error bars denote 95\% confidence intervals. Models are ordered by rollout rank. }
\label{fig:elo}
\end{figure}

We examine the bias of the real-world rollout metric itself. As shown in Fig.~\ref{fig:elo}, Elo produces more stable model orderings than partial success scores in the pour-water cross-object and fold-towel cross-environment settings.

\subsection{Correlation Uncertainty and Selection Quality}
\label{subsec:uncertainty_selection}

We quantify the correlation uncertainty caused by a limited number of rollout trials. We use paired bootstrap resampling, with the same resample used for CI-MSE and raw MSE. The 95\% confidence intervals in Table~\ref{tab:bootstrap_ci} are computed from 10,000 bootstrap resamples. Sharing resamples also enables us to calculate the improvement of CI-MSE over raw MSE. Bootstrapping does not change the best rollout-performing policy in our experiments, so CI-MSE always has zero regret@1.

\begin{table}[h]
\centering
\caption{Correlation and regret@1 confidence intervals. ``Improvement'' is the gain of CI-MSE over raw MSE.}
\label{tab:bootstrap_ci}
\small
\setlength{\tabcolsep}{3pt}
\begin{tabular}{lccc}
\toprule
Metric & Pearson $r$ [95\% CI] & Spearman $\rho$ [95\% CI] & Regret@1 [95\% CI] \\
\midrule
Raw MSE & -0.56 [-0.61, -0.48] & -0.61 [-0.67, -0.52] & 42.5 [35.4, 49.2] \\
CI-MSE & -0.74 [-0.78, -0.68] & -0.87 [-0.90, -0.80] & 0.0 [0.0, 0.0] \\
Improvement & 0.18 [0.15, 0.21] & 0.26 [0.21, 0.30] & 42.5 [35.4, 49.2] \\
\bottomrule
\end{tabular}
\end{table}

\subsection{Composition of Training and Validation Sets}
\label{subsec:data_composition}

\begin{figure}[htbp]
\centering
\includegraphics[width=0.58\linewidth]{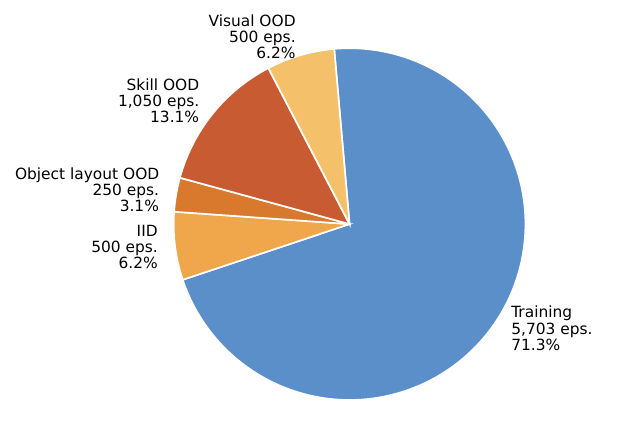}
\caption{Episode-level composition of the LBM-Eval dataset used in the simulation experiments. The dataset contains 8,003 episodes in total; labels report both episode count and percentage.}
\label{fig:data_composition_pie}
\end{figure}

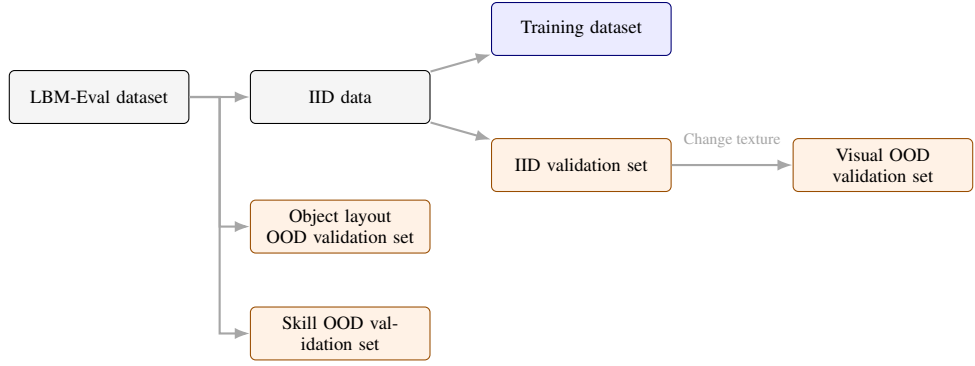
\begin{figure}[htbp]
\centering
\begin{tikzpicture}[
    font=\scriptsize,
    node distance=7mm and 8mm,
    dataset/.style={
        draw,
        rounded corners=2pt,
        align=center,
        minimum height=7mm,
        text width=22mm,
        inner sep=2.5pt,
        fill=gray!8
    },
    train/.style={dataset, fill=blue!8, draw=blue!45!black},
    val/.style={dataset, fill=orange!10, draw=orange!60!black},
    edge/.style={-{Latex[length=2mm]}, thick, draw=gray!70}
]
\node[dataset] (root) {LBM-Eval dataset};

\node[dataset, right=of root] (iiddata) {IID data};
\node[train, right=of iiddata, yshift=9mm] (train) {Training dataset};
\node[val, right=of iiddata, yshift=-9mm] (iidval) {IID validation set};
\node[val, right=16mm of iidval] (visual) {Visual OOD validation set};

\node[val, below=10mm of iiddata] (layout) {Object layout OOD validation set};
\node[val, below=of layout] (skill) {Skill OOD validation set};

\draw[edge] (root) -- (iiddata);
\draw[edge] (iiddata) -- (train);
\draw[edge] (iiddata) -- (iidval);
\draw[edge] (iidval) -- node[midway, above=1mm, font=\tiny, text=gray!70] {Change texture} (visual);
\draw[edge] (root.east) -- ++(4mm,0) |- (layout.west);
\draw[edge] (root.east) -- ++(4mm,0) |- (skill.west);
\end{tikzpicture}
\caption{Composition of the LBM-Eval training and validation sets used in the simulation experiments.}
\label{fig:data_splits}
\end{figure}

For simulation experiments, we use the LBM-Eval dataset \cite{barreiros2026careful}, which contains 49 tasks and $\sim$10k high-quality demonstrations collected through teleoperation in simulation. We split the dataset into a training set and validation sets with different distribution shifts. Figure~\ref{fig:data_composition_pie} shows the composition of the training and validation sets. Figure~\ref{fig:data_splits} shows the process of partitioning the dataset into a training set and validation sets. We first exclude 5 skills as the skill OOD validation set. The objects in the original dataset are spawned uniformly in a rectangular bounding box. We create a smaller bounding box inside the original bounding box, use episodes with objects spawned outside the smaller box as the object layout OOD validation set, and split the remaining episodes into the training set and the IID validation set. Finally, we swap the background and table texture into $15\times6$ new combinations, as shown in Figure~\ref{fig:visual_shift_textures}, and replay the episodes in the simulator to obtain the visual OOD validation set.

For real-world experiments, we use the data-scaling datasets from \cite{lin2025data}, which contain 4 tasks and 32 object-environment pairs for each task, collected using UMI. For the pour-water and arrange-mouse tasks, validation sets are provided by the authors. The cross-object validation sets contain 8 objects in the same environment. The cross-environment validation sets contain 8 environments with the same object. We use 30 episodes per object/environment for validation. For the fold-towel and unplug tasks, validation sets are not provided, so we collected an equal amount of validation data ourselves. As a result, the collectors differ between the training and validation sets, inducing an extra confounder that affects the validation error.

\subsection{Details of Rollout Evaluation Settings}

\subsubsection{Simulation Settings}
\begin{figure}[t]
\centering
\begin{minipage}[t]{0.58\linewidth}
    \centering
    \includegraphics[width=\linewidth]{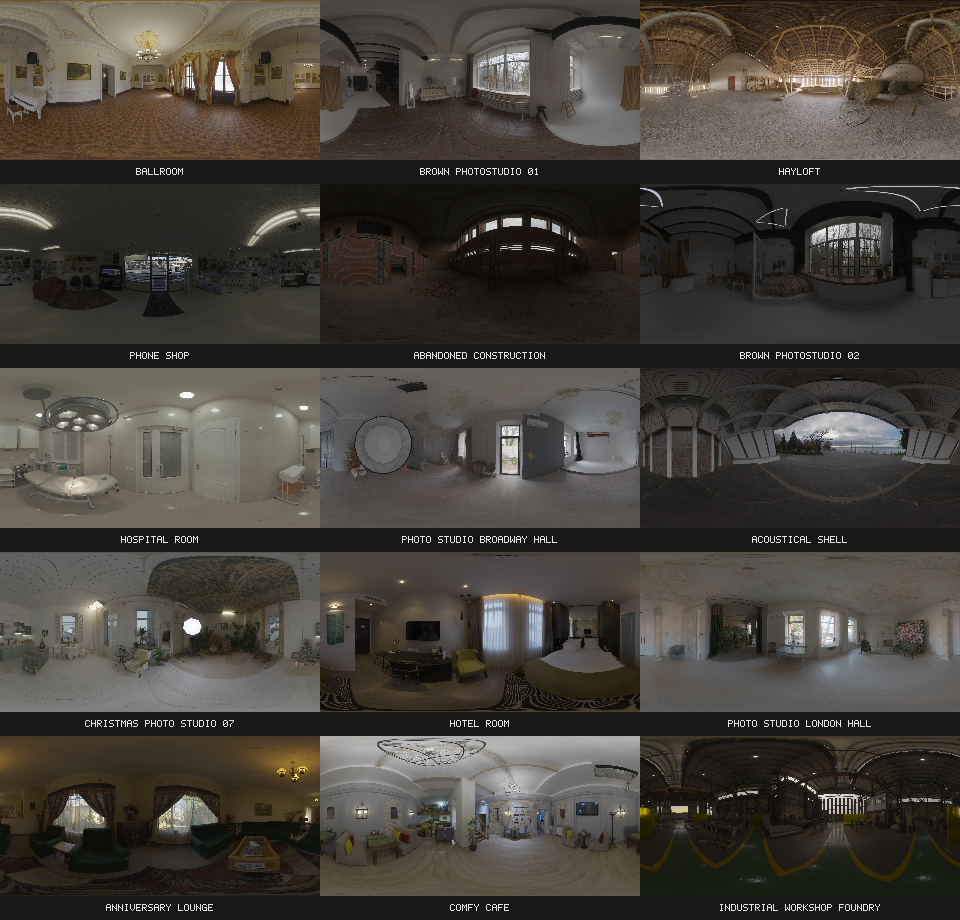}
    \vspace{1mm}
    \small (a) Environment maps
\end{minipage}
\hfill
\begin{minipage}[t]{0.38\linewidth}
    \centering
    \includegraphics[width=\linewidth]{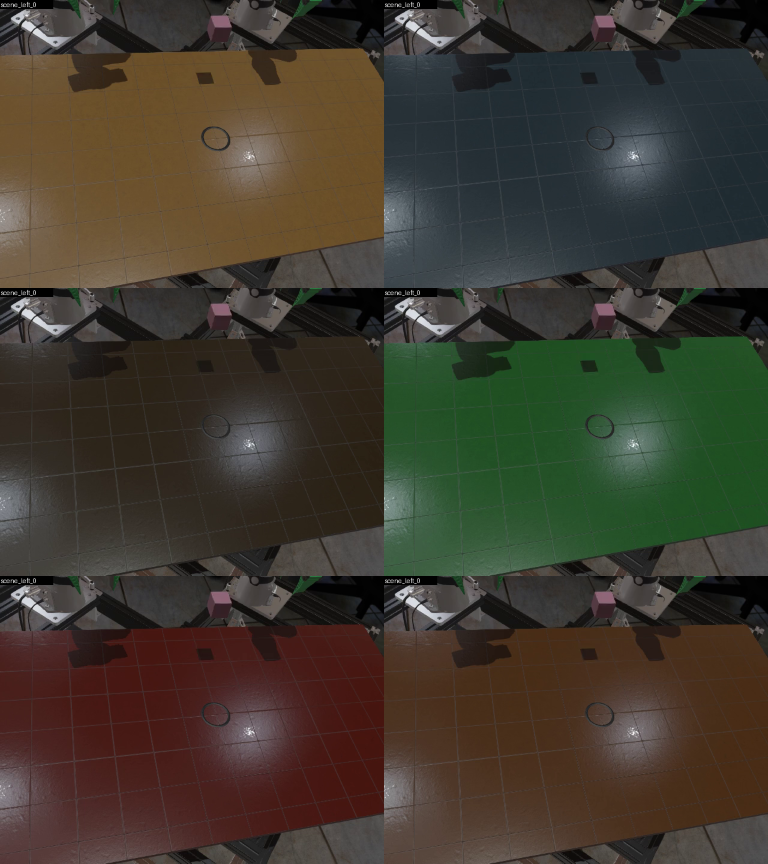}
    \vspace{1mm}
    \small (b) Table textures
\end{minipage}
\caption{Environment and table textures used in visual distribution shift evaluation.}
\label{fig:visual_shift_textures}
\end{figure}

For simulation experiments, we use the simulator and scenarios from the LBM-Eval suite. We test the policies' success rates across 12 seen tasks and 5 unseen tasks, following the validation-set splits. When evaluating under distribution shift, careful control of scenario generation is required to ensure that the distribution shift matches the validation set and differs from the training set, especially for the object layout and visual OOD settings. For object layout OOD, we use rejection sampling to generate object layouts. We sample inside the outer bounding box and reject samples in which objects are spawned inside the inner bounding box. For visual OOD, we prepare unseen environment maps and table textures, as shown in Fig.~\ref{fig:visual_shift_textures}, and then swap the background and table texture into $15\times6$ new combinations.

The main metric for measuring rollout performance is success rate. We evaluate each policy for 20 trials per skill and report the average success rate across all skills.

\subsubsection{Real-World Settings}
\label{subsec:elo_algorithm}

The real-world evaluation is conducted on a Franka arm across two categories: 8 objects and 8 environments. The trial budget is 10 trials per model per object/environment. We arrange objects according to a fixed grid layout so that different models face consistent initial states, ensuring fair pairwise comparisons.

For real-world rollout evaluation, we use Elo ranking to obtain rollout scores under limited trial budgets and noisy success rates. We convert each trial into an ordinal outcome level for pairwise comparison. These levels define an ordering only: the policy with the higher outcome level is the winner, while equal levels are ties. We do not interpret numerical gaps between adjacent levels as equal distances, and we do not use the levels as scalar partial-success scores.

All models start at $R=1500$. At each of 200,000 iterations, we uniformly sample two models, $A$ and $B$, and one rollout from each. The rubric below maps the rollouts to ordinal levels. We set $S_A=1$ if $A$ has the higher outcome level, $S_A=0$ if $B$ has the higher level, and $S_A=0.5$ for a tie; $S_B=1-S_A$. The expected scores are
\begin{equation}
E_A=\frac{1}{1+10^{(R_B-R_A)/400}},\qquad E_B=1-E_A,
\end{equation}
and the standard Elo update with $K=32$ is
\begin{equation}
R_A\leftarrow R_A+32(S_A-E_A),\qquad
R_B\leftarrow R_B+32(S_B-E_B).
\end{equation}
We repeat the sampling and update procedure for 200,000 iterations and report each model's mean rating over the latter 100,000 iterations. This procedure has no additional stability penalty or regularizer.

\noindent\textbf{Pour Water.}
\begin{quote}
\textit{Step 1: Grasping the drink bottle.}
\begin{itemize}
    \item \textbf{Level 0}: The gripper does not approach the drink bottle.
    \item \textbf{Level 1}: The gripper touches the drink bottle but does not grasp it due to minor errors, or it initially grasps the bottle but the bottle slips out during lifting.
    \item \textbf{Level 2}: The gripper pushes the drink bottle a significant distance before grasping it, or the bottle slips during the following process.
    \item \textbf{Level 3}: The gripper successfully grasps the drink bottle without any slippage.
\end{itemize}

\textit{Step 2: Pouring water into the mug.}
\begin{itemize}
    \item \textbf{Level 0}: The gripper does not approach the mug.
    \item \textbf{Level 1}: After rotating the drink bottle, its mouth remains outside the mug, making pouring impossible, or the robot knocks over the mug.
    \item \textbf{Level 2}: After rotating the drink bottle, its mouth is positioned just above the rim of the mug, allowing only partial pouring.
    \item \textbf{Level 3}: After rotating the drink bottle, its mouth is completely inside the mug, facilitating complete pouring.
\end{itemize}

\textit{Step 3: Placing the bottle on the red coaster.}
\begin{itemize}
    \item \textbf{Level 0}: The gripper does not approach the red coaster, or the bottle slips out during the moving process.
    \item \textbf{Level 1}: The drink bottle is placed outside the red coaster, the placement process disrupts the mug and causes it to topple, the robot gets stuck during rotation, the robot attempts placement but pours the water again, or the bottle is placed into the mug.
    \item \textbf{Level 2}: Only part of the drink bottle rests on the red coaster.
    \item \textbf{Level 3}: The drink bottle is fully and stably positioned on the red coaster.
\end{itemize}
\end{quote}

\noindent\textbf{Mouse Arrangement.}
\begin{quote}
\textit{Step 1: Picking up the mouse.}
\begin{itemize}
    \item \textbf{Level 0}: The gripper does not move toward the mouse or moves around it without making contact.
    \item \textbf{Level 1}: The gripper approaches the correct grasping pose and touches the mouse but drops it after lifting it slightly.
    \item \textbf{Level 2}: The gripper pushes the mouse a significant distance before grasping it, the mouse is grasped but falls when lifted higher, or the gripper grasps the mouse in an unstable way.
    \item \textbf{Level 3}: The gripper successfully grasps the mouse without any slippage.
\end{itemize}

\textit{Step 2: Placing the mouse on the mouse pad.}
\begin{itemize}
    \item \textbf{Level 0}: The gripper remains stationary in the air and fails to move toward the mouse pad, releases the mouse from a high position and causes it to fall onto the table, or drags the mouse instead of lifting it.
    \item \textbf{Level 1}: The mouse is placed outside the mouse pad, the entire mouse lands on the pad but flips because it is released from a high height, or the gripper lifts the mouse again after placing it down.
    \item \textbf{Level 2}: Only part of the mouse is placed on the mouse pad, or the entire mouse is on the pad but bounces and shifts slightly because it is released from a relatively high height or because of the gripper retraction.
    \item \textbf{Level 3}: The gripper lowers to an appropriate height before releasing the mouse, ensuring that the entire mouse is securely placed on the pad.
\end{itemize}
\end{quote}

\noindent\textbf{Fold Towel.}
\begin{quote}
\textit{Step 1: Grasping the left edge of the towel.}
\begin{itemize}
    \item \textbf{Level 0}: The gripper does not move toward the towel or moves around it without making contact.
    \item \textbf{Level 1}: The gripper moves toward the towel and attempts a grasping motion but fails to grasp any towel layer.
    \item \textbf{Level 2}: The gripper grasps only some of the towel layers, leaving others ungrasped, or grasps the towel substantially away from the edge.
    \item \textbf{Level 3}: The gripper successfully grasps all layers of the towel.
\end{itemize}

\textit{Step 2: Folding the towel to the right.}
\begin{itemize}
    \item \textbf{Level 0}: No folding motion toward the right is demonstrated.
    \item \textbf{Level 1}: After folding, the towel is left in a messy pile, such as being folded in thirds or bunched up, or the robot gets stuck during the folding process.
    \item \textbf{Level 2}: After folding, the overlap is off by more than one third, or the towel is folded in thirds with only a small top section.
    \item \textbf{Level 3}: After folding, the overlapping area exceeds two-thirds of the maximum possible overlap.
\end{itemize}
\end{quote}

\noindent\textbf{Unplug Charger.}
\begin{quote}
\textit{Step 1: Grabbing the charger.}
\begin{itemize}
    \item \textbf{Level 0}: The gripper does not grab the charger.
    \item \textbf{Level 1}: The gripper grabs the charger but not tightly enough, resulting in failure to pull out the charger.
    \item \textbf{Level 2}: The gripper securely holds the charger, but there is a collision with the power strip during the process, though the charger is eventually pulled out.
    \item \textbf{Level 3}: The gripper securely holds the charger without colliding with other objects, and the charger is successfully pulled out afterward.
\end{itemize}

\textit{Step 2: Pulling out the charger.}
\begin{itemize}
    \item \textbf{Level 0}: The charger is not pulled out.
    \item \textbf{Level 1}: The gripper holds the charger in the air, or the charger slips from the gripper after being pulled out.
    \item \textbf{Level 2}: The gripper does not place the charger far enough to the right, or the charger is released from a relatively high position.
    \item \textbf{Level 3}: The charger is successfully pulled out, and the gripper places it to the right side of the power strip.
\end{itemize}
\end{quote}

\subsection{Per-Model Rollout Scores and Validation Errors}
\label{subsec:scores}

Table~\ref{tab:appendix_scores} shows the per-model rollout success rates and offline validation errors, including both raw MSE and CI-MSE, for the simulation experiments. Table~\ref{tab:appendix_umi_scores} shows the per-model rollout scores, including Elo rankings and partial success scores, as well as offline validation errors for the real-world experiments.

\begin{table}[t]
\centering
\caption{Per-model rollout success rates and offline validation errors for the simulation experiments.}
\label{tab:appendix_scores}
\small
\setlength{\tabcolsep}{4pt}
\begin{tabular}{p{0.18\linewidth}p{0.28\linewidth}ccc}
\toprule
Variant family & Variant & \makecell{Success\\rate (\%)$\uparrow$} & \makecell{Raw MSE\\($10^{-3}$)$\downarrow$} & \makecell{CI-MSE\\($10^{-3}$)$\downarrow$} \\
\midrule
\multirow{3}{*}{Architecture} & $\pi_{0.5}$ & 53.3 & 6.690 & 2.610 \\
& X-VLA & 27.5 & 5.873 & 2.770 \\
& Gr00t N1.7 & 24.6 & 8.107 & 3.312 \\
\midrule
\multirow{5}{*}{Data scale} & 20\% & 10.8 & 5.583 & 3.046 \\
& 40\% & 17.9 & 5.686 & 2.899 \\
& 60\% & 17.5 & 5.613 & 2.908 \\
& 80\% & 20.0 & 5.909 & 2.894 \\
& 100\% & 18.8 & 6.022 & 2.914 \\
\midrule
\multirow{5}{*}{Training steps} & 20k & 7.5 & 8.508 & 3.648 \\
& 40k & 12.9 & 7.835 & 3.377 \\
& 60k & 22.5 & 6.530 & 2.934 \\
& 80k & 25.8 & 6.116 & 2.788 \\
& 100k & 27.5 & 5.873 & 2.770 \\
\midrule
\multirow{6}{*}{PEFT} & All layers, rank 8 & 3.3 & 10.091 & 4.332 \\
& All layers, rank 16 & 3.7 & 10.189 & 4.356 \\
& QKV layers, rank 8 & 10.0 & 7.893 & 3.289 \\
& QKV layers, rank 16 & 11.3 & 7.523 & 3.164 \\
& QKV+FFN layers, rank 8 & 10.8 & 7.044 & 3.193 \\
& QKV+FFN layers, rank 16 & 12.1 & 7.178 & 3.162 \\
\midrule
\multirow{5}{*}{Action head} & 175M & 17.5 & 7.938 & 3.232 \\
& 310M & 25.8 & 6.449 & 2.943 \\
& 580M & 18.8 & 6.954 & 2.905 \\
& 930M & 28.7 & 6.595 & 2.891 \\
& 1220M & 22.5 & 6.499 & 2.893 \\
\midrule
\multirow{4}{*}{VLM backbone} & florence2-base & 4.6 & 8.621 & 3.772 \\
& florence2-large & 11.3 & 7.736 & 3.159 \\
& paligemma2-3B-pt & 7.1 & 7.729 & 3.441 \\
& paligemma2-3B-mix & 10.4 & 7.390 & 3.338 \\
\bottomrule
\end{tabular}
\end{table}

\begin{table}[t]
\centering
\caption{Per-model rollout scores and offline validation errors for the real-world UMI experiments. Training pairs denotes the number of object-environment pairs used in the training data. Env. and Obj. denote cross-environment and cross-object evaluation settings.}
\label{tab:appendix_umi_scores}
\small
\setlength{\tabcolsep}{3pt}
\renewcommand{\arraystretch}{0.88}
\begin{tabular}{llccccc}
\toprule
Task & Shift & \makecell{Training\\pairs} & Elo$\uparrow$ & \makecell{Partial\\score$\uparrow$} & \makecell{Raw MSE\\($10^{-3}$)$\downarrow$} & \makecell{CI-MSE\\($10^{-3}$)$\downarrow$} \\
\midrule
Pour water & Env. & 4 & 1365.41 & 4.81 & 4.551 & 2.328 \\
 &  & 8 & 1431.11 & 5.42 & 4.264 & 2.229 \\
 &  & 16 & 1490.28 & 5.72 & 4.064 & 2.158 \\
 &  & 32 & 1713.20 & 7.22 & 4.236 & 1.991 \\
\midrule
 & Obj. & 4 & 1301.72 & 3.83 & 3.556 & 1.997 \\
 &  & 8 & 1430.44 & 5.28 & 3.063 & 1.969 \\
 &  & 16 & 1521.38 & 5.83 & 2.815 & 1.946 \\
 &  & 32 & 1746.46 & 7.38 & 2.772 & 1.911 \\
\midrule
Arrange mouse & Env. & 4 & 1456.20 & 3.98 & 1.794 & 0.470 \\
 &  & 8 & 1439.52 & 3.65 & 1.594 & 0.472 \\
 &  & 16 & 1622.17 & 5.12 & 1.404 & 0.433 \\
 &  & 32 & 1482.12 & 3.89 & 1.586 & 0.453 \\
\midrule
 & Obj. & 4 & 1479.75 & 3.11 & 1.931 & 0.613 \\
 &  & 8 & 1524.31 & 3.65 & 2.520 & 0.676 \\
 &  & 16 & 1659.96 & 4.55 & 2.628 & 0.647 \\
 &  & 32 & 1335.99 & 2.00 & 2.755 & 0.749 \\
\midrule
Fold towel & Env. & 4 & 1474.32 & 4.12 & 2.127 & 0.340 \\
 &  & 8 & 1410.74 & 4.17 & 2.346 & 0.418 \\
 &  & 16 & 1504.86 & 4.62 & 2.921 & 0.408 \\
 &  & 32 & 1610.07 & 5.10 & 2.478 & 0.396 \\
\midrule
 & Obj. & 4 & 1477.20 & 3.24 & 2.793 & 0.418 \\
 &  & 8 & 1524.19 & 3.66 & 2.694 & 0.368 \\
 &  & 16 & 1503.32 & 3.38 & 2.942 & 0.371 \\
 &  & 32 & 1495.29 & 3.42 & 2.569 & 0.412 \\
\midrule
Unplug & Env. & 4 & 1250.28 & 1.09 & 3.097 & 2.988 \\
 &  & 8 & 1534.27 & 3.66 & 3.353 & 3.330 \\
 &  & 16 & 1607.27 & 4.84 & 3.423 & 3.026 \\
 &  & 32 & 1608.18 & 4.59 & 3.228 & 3.198 \\
\midrule
 & Obj. & 4 & 1297.44 & 0.45 & 5.248 & 3.308 \\
 &  & 8 & 1495.24 & 2.52 & 5.582 & 3.382 \\
 &  & 16 & 1615.40 & 3.95 & 5.225 & 3.216 \\
 &  & 32 & 1591.92 & 4.14 & 4.625 & 2.902 \\
\bottomrule
\end{tabular}
\end{table}
\renewcommand{\arraystretch}{1.0}

\subsection{Validation-Rollout Correlation by Inference-Time Methods}

Our main experiments are carried out with temporal ensembling. Since RTC is also widely used in the community, we also study the correlation between validation error and rollout performance with RTC. We apply RTC to both real-world rollout and offline validation on the arrange-mouse task.

\begin{figure}[t]
\centering
\includegraphics[width=\linewidth]{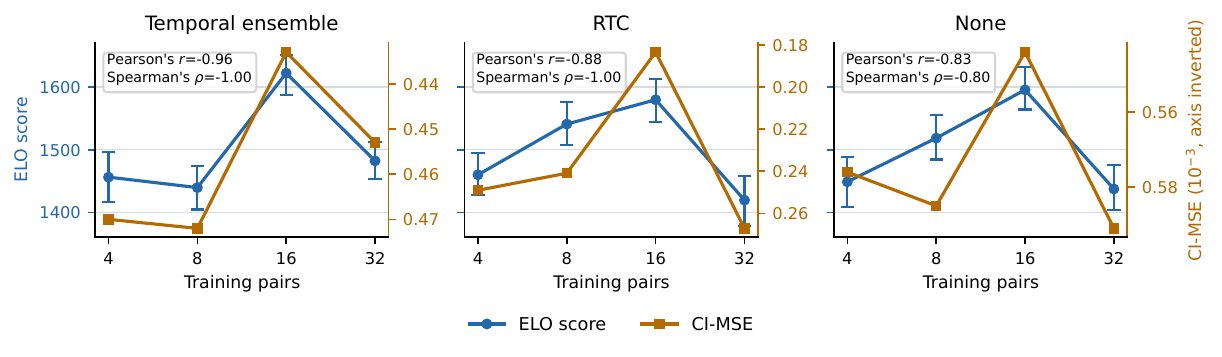}
\caption{Validation error and rollout performance under different inference-time methods on the arrange-mouse task. Blue curves show Elo scores with 95\% confidence intervals. Orange curves show CI-MSE on an inverted secondary axis to better show trend consistency with Elo scores.}
\label{fig:itm_arrange_mouse}
\end{figure}

Comparing the rollout scores of the three inference-time methods, we find that temporal ensembling produces a different ranking from the other two methods. This supports the claim that inference-time methods can materially alter policy behavior. Comparing the validation rankings with the rollout rankings, we find that CI-MSE captures differences among inference-time methods and produces consistent rankings for the temporal ensembling and RTC groups. This suggests that applying the same inference-time method to offline validation can better match rollout-time behavior.

Overall, CI-MSE achieves similar validation-rollout correlation under RTC and slightly lower correlation without any inference-time method. While this is only a single-task analysis, it suggests that the conclusions drawn under temporal ensembling are not purely an artifact of that specific inference-time method. Rather, the same validation principle appears to transfer to RTC when validation applies the corresponding rollout-time action processing. We therefore treat the temporal-ensembling results as the main experimental setting, while leaving a broader multi-task study of inference-time methods as future work.